\documentclass[]{style}

\usepackage[toc,page,header]{appendix}
\usepackage{enumitem}
\usepackage{xcolor}
\usepackage[dvipsnames]{xcolor}
\usepackage{listings}
\usepackage{caption}

\usepackage{natbib}

\usepackage{CJKutf8}

\usepackage{xargs}  

\usepackage{todonotes}  

\usepackage{multirow}

\usepackage{cleveref}

\usepackage{amsmath}
\usepackage{dsfont}

\usepackage{svg}

\usepackage{mathrsfs}
\usepackage{adjustbox}
\usepackage{multirow}
\usepackage{multicol}
\usepackage{tcolorbox}
\usepackage{changepage}
\usepackage{enumitem}
\usepackage{graphicx}
\usepackage{amssymb}
\usepackage{xcolor}
\usepackage{float}
\usepackage{multirow}
\usepackage{threeparttable}
\usepackage{graphicx}
\usepackage{subcaption}
\usepackage{algorithm}
\usepackage{algpseudocode}
\usepackage{wrapfig}
\usepackage[table]{xcolor}  
\usepackage{colortbl}       
\usepackage{tabularx} 
\usepackage{makecell} 
\usepackage[dvipsnames]{xcolor}

\newcolumntype{g}{>{\columncolor{gray!10}}c} 
\newcolumntype{A}{>{\raggedright\arraybackslash}p{4.5cm}} 
\newcolumntype{Y}{>{\centering\arraybackslash}X} 

\definecolor{catgray}{gray}{0.9}
\definecolor{skyblue}{rgb}{0.53,0.81,0.92} 

\colorlet{skyblue!30}{skyblue!30!white} 

\definecolor{customblue}{RGB}{70,130,180}  

\newtcolorbox{evolbox}[2][]{%
  enhanced,
  colframe=customblue,
  colback=white,
  coltitle=white,
  rounded corners,
  boxrule=1pt,
  titlerule=0pt,
  toptitle=1mm,
  bottomtitle=1mm,
  fonttitle=\bfseries,
  width=#2\textwidth, 
  #1
}

\usepackage{minitoc}
\usepackage{url}

\PassOptionsToPackage{table,xcdraw}{xcolor}
\usepackage{titletoc}
\usepackage{placeins}
\usepackage{pifont}

\definecolor{RowBlue}{HTML}{E9F2FB}
\definecolor{RowRed}{HTML}{F9EAEA}
\definecolor{Top1}{HTML}{50DB4B} 
\definecolor{Top2}{HTML}{A5FFA2} 
\definecolor{Top3}{HTML}{D9FFD9} 
\definecolor{Sub1}{HTML}{EAB8B8}
\definecolor{Sub2}{HTML}{E4E4E4}

\definecolor{gearred}{HTML}{D85140}
\definecolor{reprablue}{HTML}{5384ED}
\definecolor{steorange}{HTML}{EF8444}
\definecolor{softgreen}{HTML}{658E40}

\renewcommand{\emph}[1]{\textit{#1}}

\definecolor{codepink}{RGB}{220,20,120}
\definecolor{codegreen}{RGB}{0,150,0}
\definecolor{codegray}{RGB}{140,140,140}
\definecolor{codeorange}{RGB}{230,120,60}

\lstdefinestyle{pytorchstyle}{
    language=Python,
    basicstyle=\ttfamily\small,
    keywordstyle=\color{codepink}\bfseries,
    commentstyle=\color{codegray}\itshape,
    stringstyle=\color{codeorange},
    numberstyle=\tiny\color{codegray},
    numbers=none,
    showstringspaces=false,
    breaklines=true,
    frame=none,
    columns=fullflexible,
    keepspaces=true,
    xleftmargin=1.5em
}

\newcommand{\method}{JarvisHub}
\newcommand{\canvas}{\mathcal{C}}
\newcommand{\graph}{\mathcal{G}}
\newcommand{\nodes}{\mathcal{V}}
\newcommand{\edges}{\mathcal{E}}
\newcommand{\actions}{\mathcal{A}}

\newcommand{\trans}{\mathcal{F}}
\newcommand{\traj}{\tau}

\newcommand{\manifest}{\Gamma}
\newcommand{\grant}{\Omega}

\title{\method: An Open Harness for Canvas-Native Multimodal Creative Agents}

\author{JarvisX Team}

\abstract{
Creative AI is moving from single-step asset generation toward long-horizon multimodal production. Although recent generative models can synthesize high-quality images, videos, audio clips, UI elements, storyboards, slides, and other creative assets, real-world creative work requires more than isolated prompt-output interactions. It involves references, drafts, alternatives, edits, failed attempts, version relations, tool actions, evaluation signals, and human feedback, which together form an evolving project state. Existing prompt-based, chat-based, and node-based generation systems only partially support this state, as they often discard intermediate context, rely on linear conversations, or require manually specified workflows. Recent commercial systems indicate a shift toward agent-assisted creative production, but their closed architectures make it difficult to study how agents represent context, choose tools, revise artifacts, recover from failures, and maintain consistency over time. To address this gap, we introduce \method{}, a canvas-native creative agent harness for long-horizon multimodal creation. \method{} treats an editable canvas as the user workspace, the agent's external memory, action space, and shared project state, representing multimodal artifacts, dependencies, versions, and feedback as typed canvas nodes and links. Through a three-layer architecture of canvas state, protocol bridge, and agent runtime, \method{} enables agents to act within an inspectable and editable creative state. This design moves creative agents beyond isolated tool use toward sustained, human-steerable creative automation, where agents can progressively plan, generate, revise, and organize multimodal projects while users remain able to inspect, guide, and intervene throughout the process.
}

\checkdata[
\raisebox{-0.2em}{\includegraphics[width=0.025\linewidth]{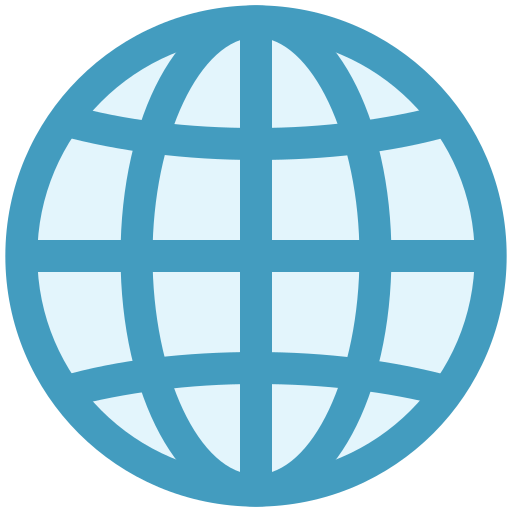}}~~Project Page]{\href{https://www.jarvishub.site/}{\texttt{https://www.jarvishub.site/}}
\\[-1.7ex]}

\checkdata[
\raisebox{-0.2em}{\includegraphics[width=0.025\linewidth]{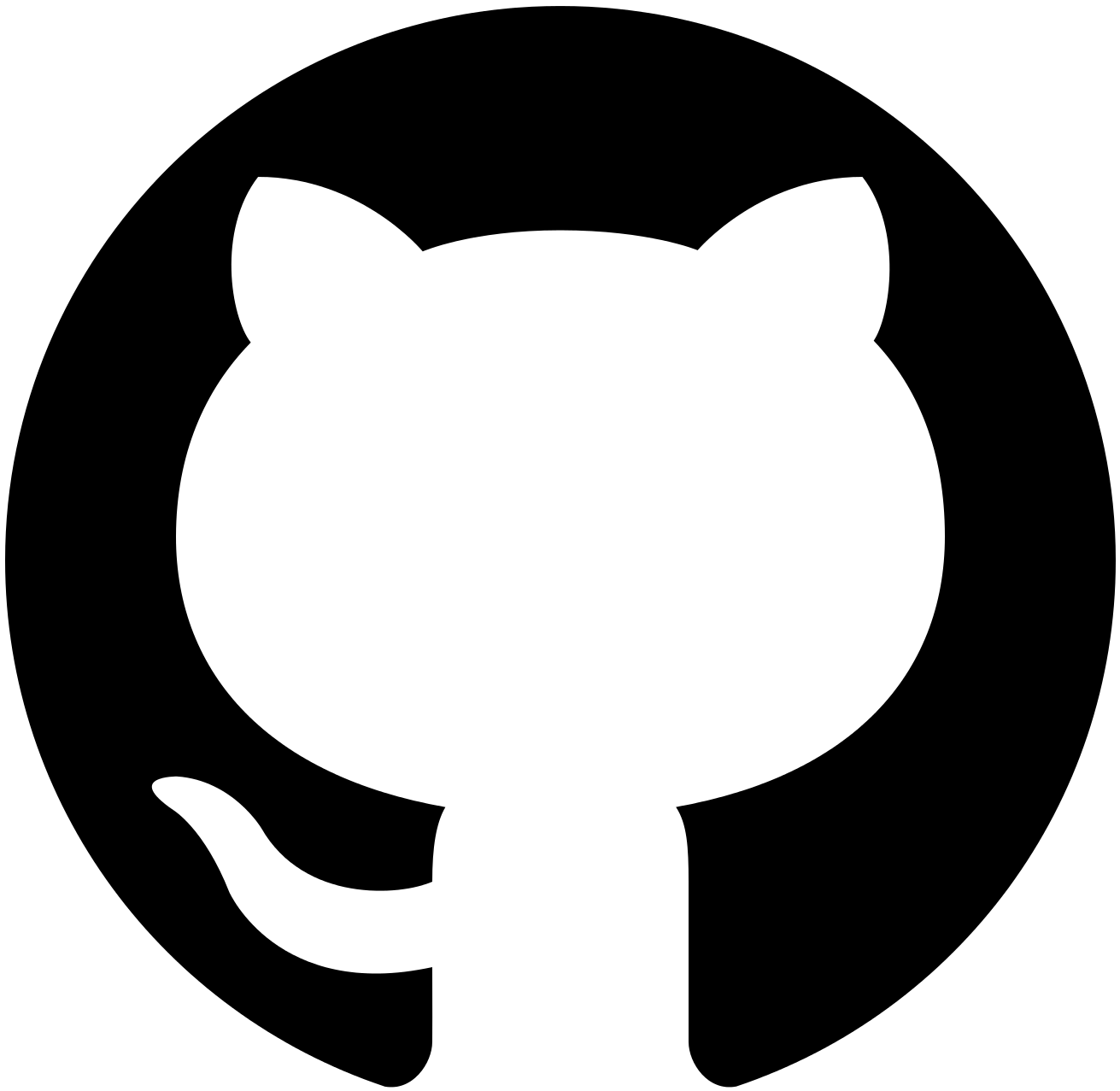}}~~GitHub Repo]{\href{https://github.com/LYL1015/JarvisHub}{\texttt{LYL1015/JarvisHub}}
\\[-1.7ex]}

\begin{document}
\maketitle

\section{Introduction}
\label{sec:intro}

Recent advances in multimodal generation have made high-quality images, videos, audio clips, UI elements, and other creative assets substantially easier to produce~\citep{seedance2,openai-gpt-image-2,si2025design2code,he2026vision2web,ding2025kimi,ge2025advancing,laurenccon2024unlocking,tian2026audio,feng2026gen,chen2026genevolve,chen2025postercraft,li2026claw}. These models are increasingly used in visual communication, UI/UX design, storyboarding, video production, slide-deck creation, and marketing content generation. Practical creative work, however, rarely follows a single prompt-output interaction. Creators typically collect references, specify styles or characters, plan layouts or shots, generate multiple candidates, revise local details, compare alternatives, incorporate feedback, and assemble intermediate results into a final deliverable. These intermediate materials—including prompts, reference images, drafts, candidates, edits, failed attempts, versions, and feedback—are not incidental by-products of creation. They form the evolving state of a creative project and provide the context needed for subsequent planning, revision, and evaluation.

This project-state view creates a concrete challenge for creative agents. They must work with an evolving workspace rather than an isolated prompt. To make a useful next move, an agent needs to know which materials already exist, how they are related, which candidates were accepted or rejected, and which parts of the project should be updated next. This context is hard to store in a plain conversation because it includes spatial layouts, version branches, references, feedback, and unfinished artifacts. The key problem is therefore not simply how to call stronger models, but how to give agents a workspace they can read, update, and keep consistent across an extended workflow.

\begin{figure}[t]
  \centering
  \includegraphics[width=\linewidth]{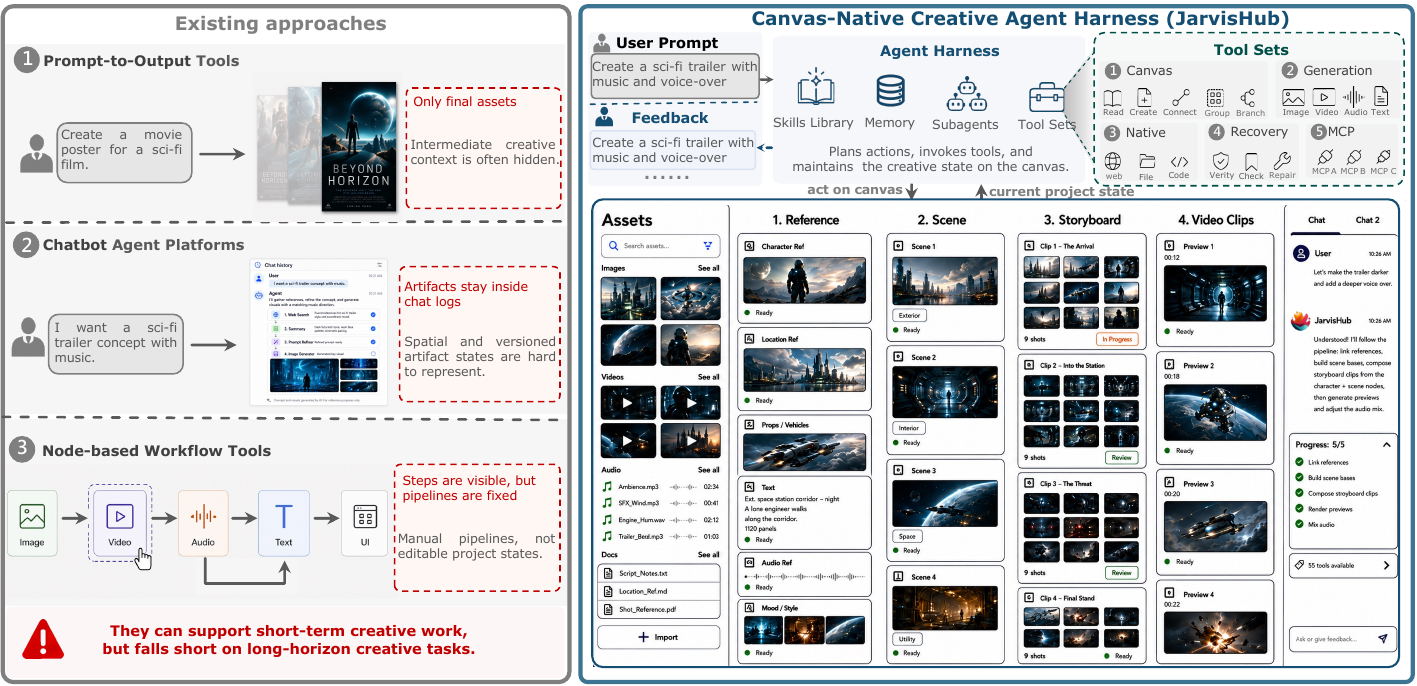}
  \caption{\textbf{Comparison of existing creative systems with JarvisHub.} Prompt-to-output tools, chatbot-based agents, and manual workflow systems each support parts of creative production, but they do not keep a unified project state for long-horizon work. JarvisHub uses the canvas as a shared workspace that agents can inspect and update, so multimodal assets, dependencies, edits, feedback, and revision traces remain accessible throughout the creative process.}
  \label{fig:introduction-system-gap}
\end{figure}

As shown in Figure~\ref{fig:introduction-system-gap}, existing systems only partially address this requirement. Prompt-to-output tools~\citep{rombach2022high,ramesh2022hierarchical,saharia2022photorealistic,yu2022scaling,brooks2023instructpix2pix,zhang2023magicbrush,zhao2024ultraedit,fu2024mgie,hui2024hq,xiao2024omnigen,liu2025step1x,wu2025qwen,seedream2025seedream,batifol2025flux,deng2025bagel,xie2025show,chen2026unify} are effective for producing individual assets, but they typically hide or discard intermediate decisions, failed trials, alternative candidates, and revision history. Chat-based creative agents~\citep{openai-gpt-image-2,google-gemini31flashimage2026,bytedance-seedream5lite2026,lin2025jarvisart,lin2025jarvisevo,chen2025photoartagent,dutt2025monetgpt} can follow multi-step instructions and call tools, but their primary context remains a linear conversation, which makes it difficult to represent spatial layouts, asset relationships, version branches, and local editing targets. Node-based workflow tools~\citep{wu2022promptchainer,xu2025comfyuicopilot,xue2025comfybench,xu2025comfyuir1,huang2025comfygpt,mozannar2025magentic,kyaw2025nodebased,kyaw2026storynodes} make execution steps more visible, but they are usually organized around manually specified pipelines rather than continuously editable project states that an agent can inspect, revise, extend, and verify. Related visual and graph-structured creative systems further explore prompt chains, branching narrative graphs, and dynamic storyboards~\citep{wu2022promptchainer,leandro2024geneva,rao2023dynamic,kyaw2025nodebased,kyaw2026storynodes}, but they remain tied to particular creative domains or manually edited graph structures rather than a general runtime harness for canvas-native agents. As a result, current systems may support useful creative actions, but they do not provide a complete runtime framework for long-horizon creative agents.

Recent commercial creative products show the same shift. Claude Design~\citep{anthropicdesign2025} supports conversational design generation and iterative refinement. Google Stitch~\citep{googlestitch2025} connects prompt-based UI generation with multi-page prototypes and front-end code. TapNow~\citep{tapnow2025} and LibTV~\citep{libtvskills2025} organize scripts, characters, storyboards, images, videos, and audio for narrative content creation. MiniMax Hub~\citep{minimaxhub2026} coordinates image, video, audio, and text generation tools in multimodal workflows. These systems suggest that creative AI is moving from isolated asset generation toward workflows where AI helps plan, revise, and organize projects across multiple stages.

However, these products remain largely closed. Researchers can observe the user-facing features, but they cannot inspect how project state is represented, how actions are checked, how tools are scheduled, how feedback is used, or how failures are repaired. This makes it difficult to study how creative agents maintain context over long workflows. The missing piece is therefore not another creative generation product, but an open harness for studying how agents operate over evolving multimodal projects.

To address this gap, we introduce \method{}, a canvas-native creative agent harness for long-horizon multimodal creation. The core idea is that the canvas is not only a visual interface for users, but also the shared project state that agents can read and update. In \method{}, prompts, references, images, videos, audio clips, UI components, storyboards, candidate results, version relations, edits, and user feedback are represented as editable canvas nodes and links. The agent observes the current canvas, interprets the available materials and their relationships, calls appropriate tools, creates or modifies nodes, and writes the results back to the same canvas. In this way, the creative process becomes visible and traceable rather than being hidden inside a sequence of prompts and tool calls.

\method{} has three key merits. First, the canvas state layer stores project materials such as nodes, attributes, layouts, versions, and dependency links. Second, the protocol bridge checks how agents read from and write to the canvas, including permissions, operation formats, validation, and logging. Third, the agent runtime chooses granted actions, calls tools, synchronizes state, and records trajectories. It exposes tool families for canvas editing, media generation, native execution, recovery, and Model Context Protocol (MCP)-backed extension. Skills, memory, and sub-agents then help organize longer workflows. Together, these layers let agents work on a shared project while preserving what they did, what evidence they used, and why the canvas changed.

Our contributions can be summarized as follows:
\begin{itemize}
\item We formalize long-horizon multimodal creation as an agent process over an editable project graph.
\item We propose and implement a canvas-native creative agent harness that unifies project state, controlled runtime execution, and trajectory recording.
\item We demonstrate \method{} on high-value long-horizon cases spanning narrative media generation, interactive web development, and presentation deck generation.
\end{itemize}


\section{Method}
\label{sec:method}

\subsection{Overview}
\begin{figure}[t]
  \centering
  \includegraphics[width=\linewidth]{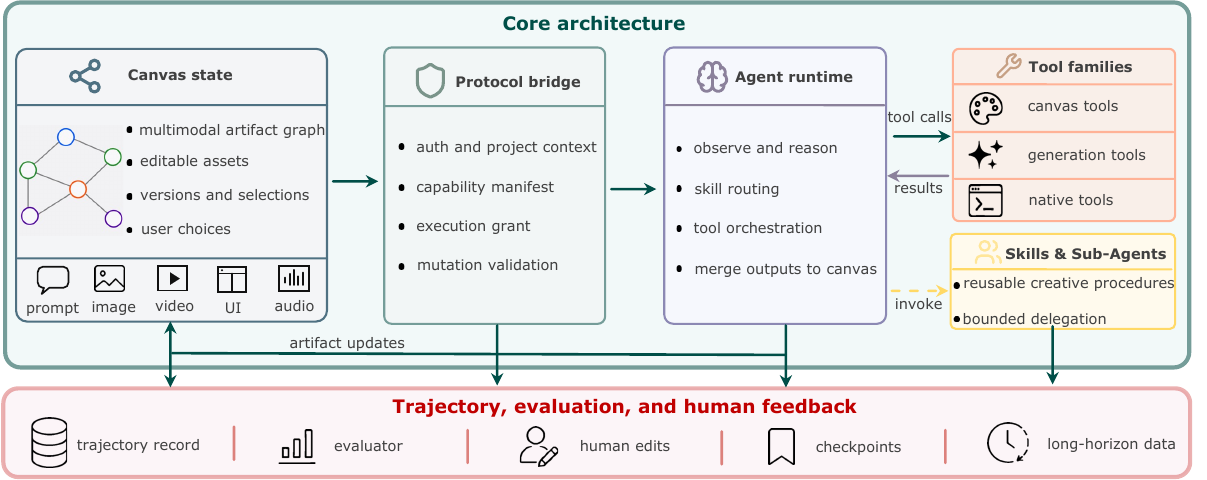}
  \caption{\textbf{Overview of the \method{} architecture.} \method{} organizes long-horizon creative work as a canvas-centered execution loop built on three core components: (1) a canvas state layer that stores multimodal artifacts, dependencies, versions, and user choices; (2) a protocol bridge that validates granted actions and controls canvas reads and writes; and (3) an agent runtime that plans, invokes tools and skills, and updates canvas artifacts. The harness further records trajectories, evaluator signals, human edits, and checkpoints as evidence for feedback, recovery, and future analysis.}
  \label{fig:architecture}
\end{figure}

\method{} is a canvas-native agent harness for long-horizon multimodal creation. It treats the canvas, not the chat history, as the main project memory. The canvas stores artifacts, dependencies, revisions, feedback, and intermediate results so that both users and agents can return to them later. As shown in Figure~\ref{fig:architecture}, each turn follows a simple loop: the runtime observes the canvas, interprets the user input, selects a permitted action through the protocol bridge, invokes the required capability, and writes the returned observation back to the canvas. Table~\ref{tab:harness-capabilities} summarizes the core capabilities supported by \method{}. The following subsections then detail how these capabilities are realized by the canvas state, the protocol bridge, the agent runtime, and the trajectory record.

\begin{table}[t]
  \centering
  \small
  \caption{Core capabilities supported by the \method{} harness.}
  \label{tab:harness-capabilities}
  \begin{tabularx}{\linewidth}{p{0.28\linewidth}X}
    \toprule
    Capability & What it enables \\
    \midrule
    Persistent project state & Keeps prompts, references, drafts, versions, outputs, and feedback in an editable canvas rather than in transient chat history. \\
    Controlled canvas action & Lets the agent read and modify the canvas only through checked operations allowed in the current turn. \\
    Tool and media execution & Connects the canvas to generation, native execution, and external tools that return inspectable artifacts. \\
    Feedback-guided revision & Uses human or model feedback to continue from accepted results, repair failed nodes locally, ask for clarification, or stop. \\
    Traceability and recovery & Records requests, actions, observations, feedback, and checkpoints so the creative process can be inspected and restored. \\
    \bottomrule
  \end{tabularx}
\end{table}

\subsection{Canvas-Native Project State}

We first define the canvas state. It is the shared project object that users edit and agents read or update. At turn $t$, we represent a creative project as:
\begin{equation}
\canvas_t = (\graph_t,\mathbf{X}_t,\mathbf{M}_t,\mathbf{U}_t,\mathbf{L}_t),
\quad
\graph_t=(\nodes_t,\edges_t).
\end{equation}
Here, $\graph_t$ is a typed artifact graph. $\nodes_t$ denotes the set of canvas nodes, and $\edges_t \subseteq \nodes_t \times \mathcal{R} \times \nodes_t$ denotes directed typed relations. Each edge can be written as $(v_i,r,v_j)$, where $r \in \mathcal{R}$ indicates a relation such as reference use, version lineage, generation dependency, grouping, or workflow continuation. The other terms store node contents and interaction records: $\mathbf{X}_t$ stores editable contents and artifact payloads, $\mathbf{M}_t$ stores provenance, execution metadata, and runtime status, $\mathbf{U}_t$ records user selections, edits, and feedback, and $\mathbf{L}_t$ stores spatial positions and group layouts. This separation lets the agent see what an artifact contains, how it was produced, how users interacted with it, and where it appears on the canvas.

We instantiate the graph through typed and addressable nodes. Let $\nodes_t=\{v_i\}_{i=1}^{N_t}$ contain $N_t$ canvas nodes. Each node is represented as:
\begin{equation}
v_i = (\mathrm{id}_i,k_i,\mathbf{p}_i,\mathbf{x}_i,\mathbf{y}_i,\mathbf{m}_i,s_i),
\quad k_i \in \mathcal{K}_{\mathrm{node}},
\end{equation}
where $\mathrm{id}_i$ is a stable identifier, $k_i$ is the node kind, $\mathbf{p}_i$ stores position and local layout, $\mathbf{x}_i$ stores editable inputs such as prompts or configuration fields, $\mathbf{y}_i$ stores generated outputs or artifact handles, $\mathbf{m}_i$ stores provenance and execution diagnostics, and $s_i$ denotes runtime status. User interaction records in $\mathbf{U}_t$ point back to the nodes, edges, or canvas regions they affect. In \method{}, the node-kind space $\mathcal{K}_{\mathrm{node}}$ covers textual, visual, audiovisual, and workflow-boundary artifacts specified by the canvas schema and capability manifest.

This representation provides three key properties for long-horizon creation. First, artifacts are addressable: the agent can refer to a specific candidate image, video clip, webpage render, or slide rather than relying on an ambiguous conversational phrase. Second, artifacts are reusable: a reference, draft, rejected candidate, or intermediate result can later serve as input to another generation or editing step. Third, dependencies are inspectable: users and agents can trace which materials influenced a result, which version was selected, and which downstream artifacts depend on it. In addition, because runtime status is part of the canvas state, later steps can distinguish completed or selected artifacts from planned, running, failed, or otherwise unverified results.

\subsection{Protocol-Constrained Canvas Interaction}

Long-horizon creation only works if canvas updates are explicit, valid, and recoverable. \method{} enforces this through a protocol bridge between the agent and the canvas. At turn $t$, the agent receives a user query $q_t$ and observes the current canvas state $\canvas_t$. The bridge provides a capability manifest $\manifest_t$, which lists the node types, mutation operations, tools, and artifact handles available in the current project. It then derives an execution grant $\grant_t$ that limits what the agent may do for the current request. Conditioned on $(q_t,\canvas_t,\manifest_t,\grant_t)$, the agent proposes an action $a_t$. The harness executes this action only if it can be encoded as a checked tool call, canvas mutation, evaluation request, clarification request, or user-facing response.

Executing a valid action produces a tool or environment observation $o_t$. The turn may also receive a feedback signal $f_t$ from a human user, a model-based critic, an auxiliary agent, or a structured evaluator. This feedback may induce a repair or follow-up decision $r_t$. The canvas then evolves as:
\begin{equation}
  \canvas_{t+1} = \trans(\canvas_t,a_t,o_t,f_t,r_t),
\end{equation}
where $\trans$ denotes the state-transition operator implemented by the bridge. It writes the accepted action and returned evidence to the current canvas, for example by creating nodes, updating fields, connecting dependencies, attaching artifacts, recording failures, creating checkpoints, or requesting human correction. Thus, canvas interaction becomes auditable rather than hidden inside language generation: if the agent modifies the canvas, the trajectory records the mutation; if it uses a generated artifact, the trajectory records both the artifact handle and the observation that produced it.

\subsection{Agent Runtime for Creative Orchestration}

\begin{figure}[t]
  \centering
  \includegraphics[width=\linewidth]{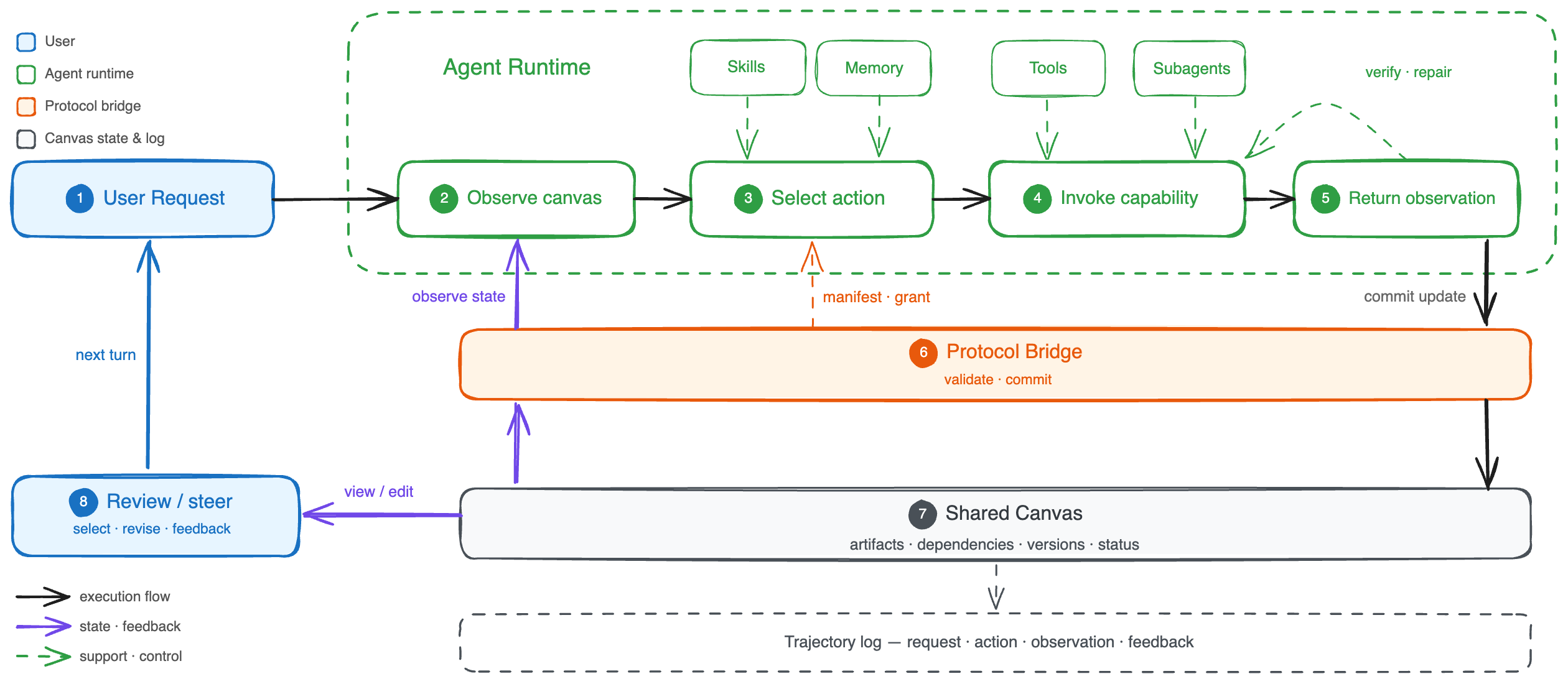}
  \caption{\textbf{Agent runtime loop.} The loop converts a user request into a protocol-checked canvas update. The runtime observes the shared canvas, selects an action under the current manifest and execution grant, invokes the corresponding capability with support from skills, memory, tools, and subagents, and returns the resulting observation to the protocol bridge. The bridge validates and commits the update to the shared canvas, where the user can inspect the result, provide feedback, and steer the next turn.}
  \label{fig:agent-runtime-loop}
\end{figure}

As illustrated in Figure~\ref{fig:agent-runtime-loop}, the agent runtime decides what the agent can do in the current turn and turns that decision into a valid canvas update. Given $(q_t,\canvas_t,\manifest_t,\grant_t)$, it first interprets the user request against the current canvas. It then filters the available operations by the execution grant, invokes the required capability, and returns the resulting evidence through the protocol bridge. This gives each turn a granted action space:
\begin{equation}
  \actions_t = \actions(\grant_t,\manifest_t,\canvas_t,q_t),
  \quad a_t \in \actions_t.
\end{equation}
Here, $\actions_t$ denotes the actions that are both relevant to the request and permitted in the current turn. The formula is not a list of implementation tools. Instead, it states the runtime contract: an accepted action must be grounded in the observed canvas, exposed by the capability manifest, allowed by $\grant_t$, and committed through the protocol bridge. This keeps the shared canvas as the project state, rather than letting the runtime maintain a separate hidden state.

To implement this contract, the runtime groups available capabilities into a small set of tool families. Each family can be granted for a turn, invoked by the agent, and committed back to the canvas through the same bridge. Table~\ref{tab:runtime-capabilities} summarizes these families.

\begin{table}[t]
  \centering
  \small
  \caption{Core tool families used by the agent runtime.}
  \label{tab:runtime-capabilities}
  \begin{tabularx}{\linewidth}{p{0.25\linewidth}p{0.22\linewidth}X}
    \toprule
    Tool family & Runtime role & Representative operations \\
    \midrule
    Canvas tools & Update project state. & Read, create, update, connect, group, select, and branch nodes; maintain artifact handles and runtime status. \\
    Generation tools & Produce canvas artifacts. & Image, video, audio, and composed-media generation. \\
    Native tools & Use external execution. & Browser, file, code, search, document, and presentation operations that return inspectable artifacts. \\
    Recovery tools & Inspect and repair. & Structured feedback, verification, checkpointing, and local repair. \\
    MCP tools & Extend external services. & MCP-provided capabilities under the same manifest-and-grant contract. \\
    \bottomrule
  \end{tabularx}
\end{table}

The tool families define what can be invoked. The runtime also uses three higher-level supports to decide how those capabilities should be sequenced across long creative tasks:
\begin{itemize}[leftmargin=*]
  \item \textbf{Skills.} Skills encode reusable creative procedures, such as storyboarding, reference-guided generation, design-to-web reconstruction, video prompting, or deck construction. They give the runtime a task structure without hard-coding every step into the bridge.
  \item \textbf{Memory.} Memory preserves user preferences, prior decisions, and procedural knowledge across turns. It helps the runtime choose actions that remain consistent with earlier project context.
  \item \textbf{Subagents.} Subagents handle independent subtasks when a workflow can branch. For example, separate video shots can be explored in parallel, while the parent agent still selects useful results and integrates them back into the shared canvas graph.
\end{itemize}
Together, the tool families provide executable capabilities, while skills, memory, and subagents help choose and sequence them. The resulting workflow can explore alternatives, but its state remains anchored in the canvas, execution grants, and protocol-checked updates.

\subsection{Feedback and Traceability}

Long-horizon creative work requires feedback before the final artifact is complete. A feedback signal $f_t$ may come from a user, an evaluator, or a subagent critic. Users can select candidates, reject versions, revise prompts, adjust styles, or identify local defects, while JarvisHub can automatically inspect node outputs for consistency, missing evidence, visual quality, and task-specific constraints. The purpose of feedback is not merely to judge an intermediate result, but to determine the next state transition: whether the agent should continue from an accepted node, repair a failed node locally, ask the user for clarification, or stop when the available evidence is insufficient.

To make feedback actionable and traceable, the harness records each feedback signal together with the state and execution context in which it occurs. We formally define the trajectory as:
\begin{equation}
  \traj =
  \left\{
    (q_t,\canvas_t,\manifest_t,\grant_t,a_t,o_t,f_t,r_t,\canvas_{t+1})
  \right\}_{t=1}^{T},
\end{equation}
where, $q_t$ is the user request, $\canvas_t$ is the canvas state before execution, $\manifest_t$ specifies the available capabilities, $\grant_t$ constrains the actions permitted in the current turn, $a_t$ is the selected agent action, $o_t$ is the tool observation or generated evidence, $f_t$ is the feedback signal, $r_t$ is the repair or follow-up decision induced by the feedback, and $\canvas_{t+1}$ is the updated canvas state.

\section{Experiments}
\label{sec:experiments}

We evaluate \method{} on representative long-horizon creative tasks to demonstrate its capability in realistic creative workflows. Specifically, the experiments examine whether a canvas-native harness can support high-value tasks that require persistent project context, iterative artifact generation, and feedback-driven refinement.

\subsection{Experimental Setup}

All tasks are run in the same \method{} environment. We use GPT-5.5~\citep{openai-gpt-5.5} as the main agent backend, GPT Image 2~\citep{openai-gpt-image-2} as the image-generation backend, Seedance 2.0~\citep{seedance2} as the video-generation backend, and Gemini 3.1 Pro~\citep{gemini-3.1-pro} as the multimodal evaluation backend. More details about the model-backend configuration are provided in our open-source GitHub repository: \url{https://github.com/LYL1015/JarvisHub}.

\subsection{Long-Horizon Creative Tasks}

As shown in Table~\ref{tab:creative-tasks}, we evaluate \method{} on three representative long-horizon creative tasks: narrative media generation, interactive web development, and presentation deck generation. These tasks cover common creative workflows where an open-ended goal must be developed into a coherent deliverable. They require planning, reference organization, intermediate artifact generation, feedback, and local revision.

\begin{table}[t]
\centering
\small
\caption{High-value long-horizon creative tasks used in our experiments.}
\label{tab:creative-tasks}
\begin{tabularx}{\linewidth}{>{\raggedright\arraybackslash}p{0.27\linewidth}>{\raggedright\arraybackslash}p{0.38\linewidth}>{\raggedright\arraybackslash}X}
\toprule
Task & Definition and challenge & Representative outputs \\
\midrule
Narrative media generation &
Convert a story, script, or scene brief into a temporally coherent visual or audiovisual sequence, testing narrative planning, identity preservation, style consistency, and cross-shot continuity. &
Character references, scene designs, storyboard panels, shot plans, generated image sequences, video clips, and animatics. \\
Interactive web development &
Convert a design goal, information structure, or interaction requirement into a rendered web artifact, testing coordination among layout design, interaction logic, frontend code, preview inspection, and iterative revision. &
Static pages, dynamic websites, landing pages, web interfaces, interactive prototypes, rendered previews, and frontend implementati ons. \\
Presentation deck generation &
Convert a topic, source document, report, or communication goal into a coherent multi-slide presentation, testing content selection, narrative organization, slide layout, visual synthesis, and cross-slide consistency. &
Presentation decks, academic talks, project reports, pitch decks, interview presentations, visual summaries, and explanatory diagrams. \\
\bottomrule
\end{tabularx}
\end{table}

\subsection{Qualitative Results}

For each task, we report two complementary views: a workspace trace of the canvas-centered production process and a generated artifact of the final deliverable. These paired views show how plans, references, assets, dependencies, and agent progress remain inspectable throughout long-horizon work, and how accumulated canvas state supports coherent outputs.

\begin{figure}[H]
\centering
\setlength{\abovecaptionskip}{0.1cm} 
\includegraphics[width=0.95\linewidth]{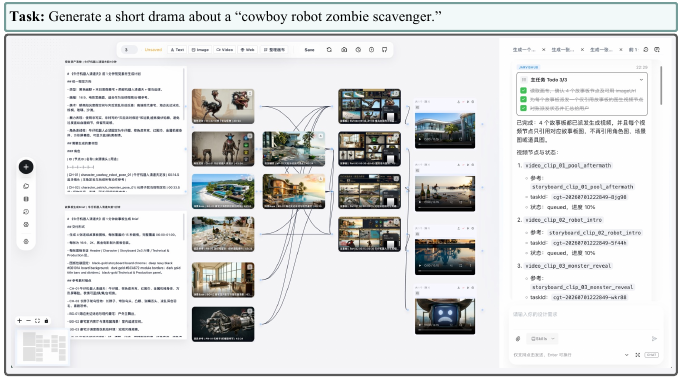}
\caption{\textbf{Workspace trace for narrative media generation.} The \method{} canvas externalizes the task brief, planning notes, visual references, shot candidates, dependency links, and agent progress for a short-drama case.}
\label{fig:qualitative-narrative-media-generation-workspace}
\vspace{-0.6cm}
\end{figure}

\begin{figure}[H]
\centering
\setlength{\abovecaptionskip}{-0.2cm} 
\includegraphics[width=0.95\linewidth]{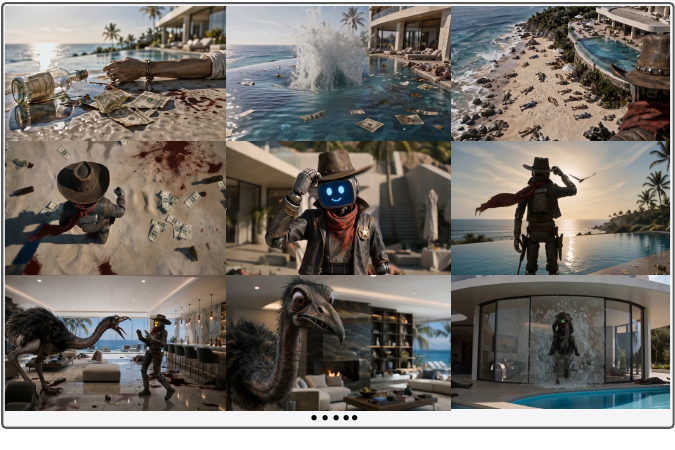}
\caption{\textbf{Generated artifact for narrative media generation.} The final output presents key frames produced through the canvas-managed workflow, with recurring characters, setting cues, and action continuity across shots. This case is inspired by \textbf{Mx-Shell}'s original work \textit{Zombie Sweeper}.}
\label{fig:qualitative-narrative-media-generation-artifact}
\end{figure}

\paragraph{Narrative media generation.}
As shown in Figures~\ref{fig:qualitative-narrative-media-generation-workspace} and~\ref{fig:qualitative-narrative-media-generation-artifact}, the narrative media case turns a short-drama prompt into a coherent visual sequence. The workspace trace makes story planning, visual references, dependencies, and generation progress inspectable. The output illustrates how the canvas preserves narrative and visual continuity across shots.

\begin{figure}[H]
\centering
\setlength{\abovecaptionskip}{0.1cm} 
\includegraphics[width=0.92\linewidth]{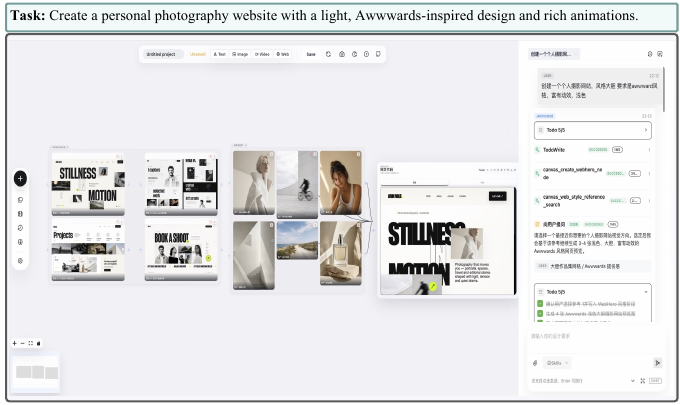}
\caption{\textbf{Workspace trace for interactive web development.} The \method{} canvas tracks the web brief, visual references, layout drafts, implementation artifacts, previews, and revision state during website construction.}
\label{fig:qualitative-web-development-workspace}
\vspace{-0.7cm}
\end{figure}

\begin{figure}[H]
\centering
\setlength{\abovecaptionskip}{0.1cm} 
\includegraphics[width=0.92\linewidth]{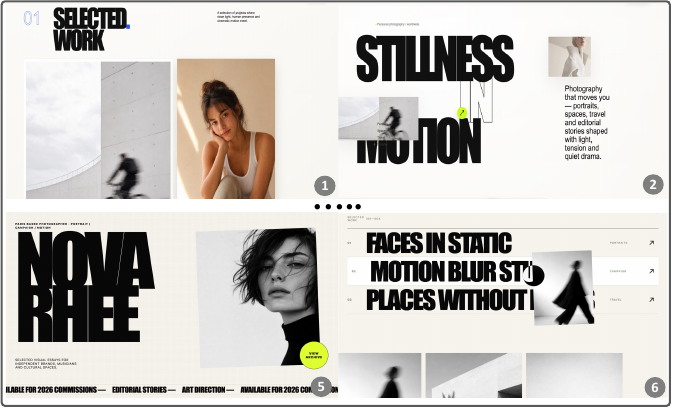}
\caption{\textbf{Generated artifact for interactive web development.} The final output shows website screens with consistent typography, image placement, page structure, and visual direction.}
\label{fig:qualitative-web-development-artifact}
\end{figure}

\paragraph{Interactive web development.}
As shown in Figures~\ref{fig:qualitative-web-development-workspace} and~\ref{fig:qualitative-web-development-artifact}, the web development case turns an aesthetic and interaction brief into a rendered photography website. The workspace trace makes design references, implementation progress, previews, and revision state inspectable. The output illustrates how the canvas preserves visual direction and interface consistency during iterative web construction.

\begin{figure}[H]
\centering
\setlength{\abovecaptionskip}{0.1cm} 
\includegraphics[width=0.92\linewidth]{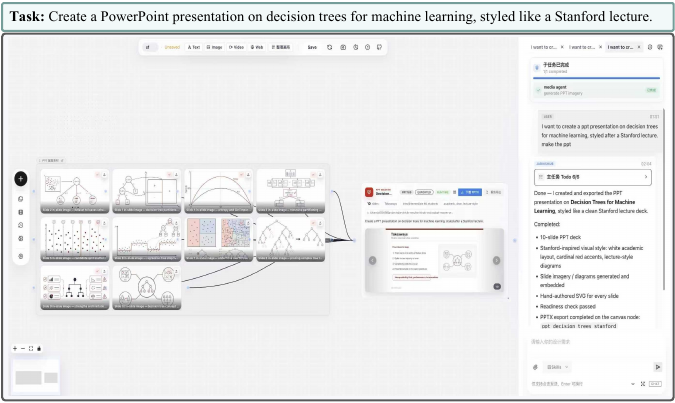}
\caption{\textbf{Workspace trace for presentation deck generation.} The \method{} canvas records lecture content, generated diagrams, slide drafts, dependency links, PowerPoint previews, and revision state.}
\label{fig:qualitative-presentation-deck-generation-workspace}
\vspace{-0.4cm}
\end{figure}

\begin{figure}[H]
\centering
\setlength{\abovecaptionskip}{0.1cm} 
\includegraphics[width=0.92\linewidth]{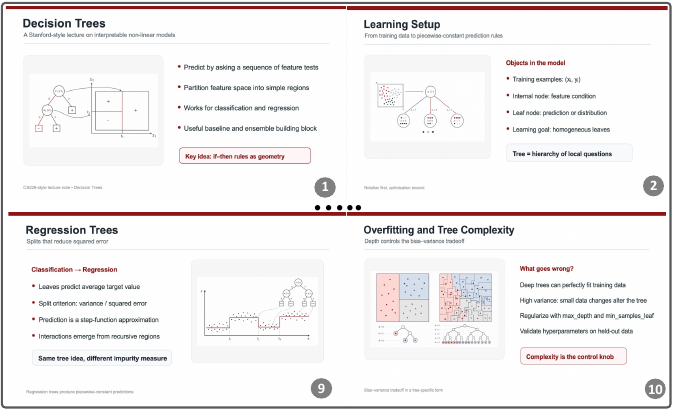}
\caption{\textbf{Generated artifact for presentation deck generation.} The final output shows lecture slides with consistent layout, diagram style, emphasis color, and slide-level organization.}
\label{fig:qualitative-presentation-deck-generation-artifact}
\end{figure}

\paragraph{Presentation deck generation.}
As shown in Figures~\ref{fig:qualitative-presentation-deck-generation-workspace} and~\ref{fig:qualitative-presentation-deck-generation-artifact}, the presentation case turns a machine-learning lecture topic into a structured slide deck. The workspace trace makes content planning, visual assembly, previews, and task progress inspectable. The output illustrates how the canvas preserves topic structure and visual style across a multi-slide deliverable.

\section{Discussion}
\label{sec:discussion}

\paragraph{Why should the canvas be treated as an agent workspace?}
The canvas is useful not only because it gives users a visual interface, but because it can serve as a shared workspace for both humans and agents. In \method{}, prompts, references, candidates, edits, versions, dependencies, and feedback are represented as typed and addressable canvas nodes and links. This turns the canvas into both an external memory and an action space for the agent. Users can inspect and guide the same state that the agent reads and modifies. As a result, the agent can reuse prior artifacts, perform local updates, maintain dependencies, and continue unfinished work without hiding the process in private tool calls or transient chat history.

\paragraph{What can an open harness enable for future creative-agent research?}
An open creative-agent harness can support research artifacts beyond a single system demonstration. First, it enables project-state benchmarks, where each task specifies an initial canvas, reference materials, available tools, constraints, feedback events, and expected checkpoints rather than only a prompt and a target answer. Second, it supports evaluation protocols that combine final artifact quality with process-level measures, such as context preservation, tool-use appropriateness, dependency correctness, feedback adherence, and repair success. Third, it creates a data flywheel for improving future models: each run produces structured examples of canvas states, agent actions, tool observations, feedback signals, repair decisions, and final outcomes. With proper consent, anonymization, copyright filtering, and quality control, these trajectories can be used to train future creative agents for planning, tool selection, multimodal state tracking, local repair, and feedback-guided revision.

\paragraph{Why should trajectories be the object of analysis?}
For creative tasks, the final artifact alone does not fully characterize agent behavior. Multiple outputs may be acceptable, and a visually strong result may still be produced by a brittle or incorrect process. Conversely, an imperfect final result may still contain useful intermediate artifacts and recoverable decisions. The trajectory records how the agent moves from one canvas state to the next: what context it uses, which action it selects, what tool evidence is returned, what feedback is received, and how the canvas is repaired or updated. This makes process-level failures visible, such as ignoring references, losing accepted choices, overwriting useful variants, or regenerating globally when a local repair is needed.

\section{Conclusion and Limitations}
\label{sec:conclusion}

We presented \method{}, a canvas-native agent harness for long-horizon multimodal creation. \method{} represents a creative project as an editable canvas graph. Through a protocol bridge, agents can inspect project context, call tools, update canvas artifacts, incorporate feedback, and record state changes. The runtime integrates canvas operations, media generation, native tools, repair, checkpointing, and trajectory capture in one shared workspace. Experiments on representative creative tasks show that \method{} can support more inspectable and reusable agentic creative workflows. The recorded trajectories also provide useful data for building benchmarks, evaluating long-horizon behavior, and training future creative agents.

Several limitations remain. First, our experiments are qualitative demonstrations rather than a completed benchmark or leaderboard. Second, \method{} focuses on orchestration and project-state management, so final artifact quality still depends on the external models and tools used by the runtime. Third, the protocol bridge makes canvas actions explicit and recoverable, but it does not guarantee that the agent's creative decisions are semantically correct. Finally, trajectory records are valuable for analysis and training, but raw trajectories require quality filtering, consent, anonymization, and copyright review before they can be used as research data.

\section{Contributions}
\label{sec:contributions}

The contributors' names are listed in alphabetical order (A to Z) by first name.

\noindent\textbf{Core Contributors}\\
Yunlong Lin, Zixu Lin, Zhaohu Xing

\vspace{1em}
\noindent\textbf{Contributors}\\
Biqiang Li, Chenxin Li, Haonan Wang, Haitao Wu, Hengyu Liu, Jianghai Chen, Kaituo Feng, Kaixin Li, Shawn Chen, Shijue Huang, Sixiang Chen, Tsung-Yi Ho, Wenxuan Huang, Xiangyan Liu, Xiaomeng Hu, Xuanhua He, Yan Sun, Yunqing Zhao, Zhiqin Yang, Zehan Wang, Zhengyang Tang

\vspace{1em}
\noindent\textbf{Academic Advisors}\\
Tianyu Pang, Xiangyu Yue

\clearpage

\bibliographystyle{plainnat}
\setlength{\bibhang}{0pt}
\setlength\bibindent{0pt}
\bibliography{main}


\end{document}